\documentclass[12pt]{iopart}

\usepackage{amssymb}
\usepackage{graphicx}
\usepackage{tabularx}
\usepackage{array}
\usepackage{multicol}
\usepackage{comment}
\usepackage{xcolor}
\usepackage{soul}
\usepackage{cite}
\usepackage{threeparttable}

\newcommand{\argmax}{\mathop{\rm argmax}\limits}
\newcommand{\argmin}{\mathop{\rm argmin}\limits}

\newcolumntype{R}[1]{>{\raggedleft\arraybackslash}p{#1}}
\newcolumntype{C}[1]{>{\centering\arraybackslash}p{#1}}
\newcolumntype{L}[1]{>{\raggedright\arraybackslash}p{#1}}

\graphicspath{ {./images/} }

\begin{document}

\title{Boosting Template-based SSVEP Decoding by Cross-domain Transfer Learning}
\author{Kuan-Jung Chiang$^{\dag,\S}$, Chun-Shu Wei$^{\ddag}$, Masaki Nakanishi$^{\S}$, and Tzyy-Ping Jung$^{\S,\ddag}$}
\ead{kuchiang@eng.ucsd.edu}
\address{$^{\dag}$ Department of Computer Science and Engineering, University of California San Diego}
\address{$^{\ddag}$ Department of Computer Science, National Chiao Tung University}
\address{$^{\S}$ Institute for Neural Computation and Institute of Engineering in Medicine, University of California San Diego}

\vspace{10pt}

\begin{abstract}
\textit{Objective}:
This study aims to establish a generalized transfer-learning framework for boosting the performance of steady-state visual evoked potential (SSVEP)-based brain-computer interfaces (BCIs) by leveraging cross-domain data transferring.
\textit{Approach}:
 We enhanced the state-of-the-art template-based SSVEP decoding through incorporating a least-squares transformation (LST)-based transfer learning to leverage calibration data across multiple domains (sessions, subjects, and EEG montages).
\textit{Main results}:
Study results verified the efficacy of LST in obviating the variability of SSVEPs when transferring existing data across domains. 
Furthermore, the LST-based method achieved significantly higher SSVEP-decoding accuracy than the standard task-related component analysis (TRCA)-based method and the non-LST naive transfer-learning method.
\textit{Significance}:
This study demonstrated the capability of the LST-based transfer learning to leverage existing data across subjects and/or devices with an in-depth investigation of its rationale and behavior in various circumstances. The proposed framework significantly improved the SSVEP decoding accuracy over the standard TRCA approach when calibration data are limited. Its performance in calibration reduction could facilitate plug-and-play SSVEP-based BCIs and further practical applications.

\end{abstract}

%
%
%
%
%

\section{Introduction}
A brain-computer interface (BCI) is an immediate pathway that provides an intuitive interface for users, especially disabled ones, to translate their intention into commands to control external devices \cite{wolpaw2002brain}. Among various neuroimaging modalities,  electroencephalogram (EEG) has been widely used for developing BCI applications for the sake of unobtrusiveness, low cost, and high temporal resolution \cite{wolpaw2002brain}. Researchers have attempted to develop a variety of real-world applications of an EEG-based BCI such as spellers that allow disabled users to communicate with others. Farewell and Donchin demonstrated a P300-based BCI as the first-ever BCI speller in the 1980s \cite{donchin:1988}. Since then, BCI spellers based on visual-evoked brain responses have been improved by different advancements in the decoding techniques. Recently, steady-state visual evoked potentials (SSVEPs), a type of  neural responses to repetitive visual stimulation, have attracted increasing attention and been used in high-speed BCI spellers as the acquisition of SSVEP is relatively stable \cite{wang2008brain, vialatte:2010, gao2014visual}. The performance of SSVEP-based BCIs has been significantly improved by advances in system design, signal processing, and decoding algorithms in the past decade \cite{chen2015high}.

As SSVEPs are oscillatory fluctuations that respond to flickering visual stimuli, an SSVEP-based BCI measures the user's SSVEP and uses an algorithm to identify the corresponding stimulus based on the SSVEP data. Analyzing SSVEPs in the frequency-domain has been an intuitive approach to detect the frequency peaks in the spectral response of SSVEP that correspond to the flickering frequency of stimuli. For example, power spectral density analysis (PSDA) \cite{wang2008brain} that uses features in frequency-domain was proposed to decode SSVEPs. Canonical correlation analysis (CCA) \cite{lin2006frequency, bin:2009b} that compares a test trial with computer-generated SSVEP models/templates consisting of sine-cosine signals for each target \cite{wang2015computational} was also proved to work on SSVEPs. However, these methods failed to maintain consistent performance across users because the individual differences in the SSVEP data were not taken into account \cite{wei2018subject}. Later studies exploited individualized training data, or SSVEP templates, and developed training-based SSVEP decoding schemes that better characterize individual SSVEP patterns using personalized calibration data. The training-based methods usually outperform the aforementioned calibration-free methods \cite{nakanishi2014high, nakanishi2017enhancing, nakanishi2015comparison, zhang:2013l1regularized, zhang2014frequency, zerafa2018to}.


The success of training-based SSVEP decoding has significantly boosted the efficiency of SSVEP-based BCI spellers. Nonetheless, the calibration procedure is often laborious and time-consuming, hindering practical and wide-spread applications of BCIs in our daily life. Several studies have attempted to adopt transfer-learning techniques to reduce the individual-variability or session-variability so that a model for SSVEP can be tuned with existing data without repeat collections of calibration data before each use \cite{Wu2020transfer}. For instance, Yuan \etal and Wong \etal proposed subject-to-subject transfer-learning methods that transfer SSVEP data from existing subjects to new ones using a spatial filtering approach \cite{yuan2015enhancing, wong2020inter}. Rodrigues \etal also proposed a cross-subject transfer learning method using a Riemannian geometrical transformation \cite{Rodrigues2019-yl}. Waytowich \etal applied a convolutional neural network to train a subject-independent classifier for detecting SSVEPs \cite{Waytowich2018-ph}. In another study, Nakanishi \etal proposed a session-to-session and device-to-device transfer-learning method using spatial filtering \cite{Nakanishi2016-qa, nakanishi2019facilitating}. Suefusa \etal adopted a frequency shifting technique to synthesize calibration data at arbitrary frequencies from existing calibration data \cite{Suefusa2017-qp}.
  
  The transfer learning approaches have succeeded either in improving classification accuracy of SSVEPs compared with calibration-free methods without any calibration process or compared with fully-calibrated methods with reduced calibration cost. However, such methods focused only on transferring data across one domain such as cross-subject or cross-session transferring. For example, the aforementioned subject-to-subject transfer learning methods implicitly assumed that their subjects, stimulus parameters, and EEG devices are identical between calibration and target data \cite{yuan2015enhancing, wong2020inter, Rodrigues2019-yl, Waytowich2018-ph}. Similarly, the cross-session and the cross-device transfer learning methods proposed in our previous studies implicitly required calibration data to be from the same subjects and stimulus parameters with target data \cite{Nakanishi2016-qa, nakanishi2019facilitating}. Table \ref{table_compare} compares and summarizes current transfer-learning schemes for SSVEP-based BCIs. It is worth noting that a generalized transfer learning approach that can handle transferring existing data across multiple domains has yet to be proposed.
  
    \fulltable{\label{table_compare}Comparison between different transfer learning methods for SSVEP-based BCIs.}
  \begin{threeparttable}
      \begin{tabular}{llllcccc}
         \br
         Study & Year & Calibration & Transferring & \multicolumn{4}{c}{Domain transferred}   \\ \cline{5-8} 
          & & data \tnote{*1} & approach & Subjects & Sessions & Devices & Stimuli  \\
         \mr
         \cite{yuan2015enhancing} & 2015 & Not required & Spatial filtering & \textbf{Yes} & No & No & No  \\
         \cite{Nakanishi2016-qa} & 2016 & Not required & Spatial filtering & No & \textbf{Yes} & No & No \\
         \cite{Suefusa2017-qp} & 2017 & Not required & Frequency shifting & No & No & No & \textbf{Yes} \\ 
         \cite{Waytowich2018-ph} & 2018 & Not required & Convolutional & \textbf{Yes} & No & No & No \\
         & & & neural network & & & & \\
         \cite{Rodrigues2019-yl} & 2019 & Not required & Geometrical& \textbf{Yes} & No & No & No \\
         & & & transformation & & & & \\
         \cite{nakanishi2019facilitating} & 2019 & Not required & Spatial filtering & No & No & \textbf{Yes} & No  \\        
        \cite{chiang2019cross} & 2019 & Required & Channel-wise & \textbf{Yes} & No & No & No \\
        &&& projection &&&&\\
        \cite{wong2020inter} & 2020 & Required & Spatial filtering & \textbf{Yes} & No & No & No  \\
        \br
    \end{tabular}
    \begin{tablenotes}\footnotesize
    \item[*1] Additional calibration data from a target session.
    \end{tablenotes}
    \end{threeparttable}
  \endfulltable

The goal of this study is to propose a generalized framework of transfer learning that can leverage SSVEP data across multiple domains including subjects, sessions, and devices toward a practical BCI application. Our preliminary study demonstrated that applying LST to the existing SSVEP datasets acquired from other users can augment the size of calibration data for new users and in turn enhance decoding performance without acquiring extra calibration process \cite{chiang2019cross}.
However, although the LST-based method could technically be employed to transfer SSVEPs across any domain such as cross-session and cross-device transferring, the effectiveness of such scenarios has not been investigated in our previous work \cite{chiang2019cross}.
Therefore, the generalizability of the LST-based method, especially in a real-world scenario where EEG data from different users are more likely recorded by different types of EEG recording systems, remains unknown. The discrepancy among recording montages has posed a grand challenge in consolidating EEG data across domains and hindered the translation of BCI technologies toward practical applications.

This study investigated the feasibility and the effectiveness of the LST-based method in transferring SSVEPs across multiple domains using the dataset acquired from multiple subjects and sessions with multiple EEG devices collected in our previous study \cite{nakanishi2019facilitating}. This work further examined and compared the characteristics of data from different subjects with and without the LST-based transferring in feature spaces.
In addition, this study also investigated the effects of parameters such as the number of data transferred via the LST on the classification accuracy.


\section{Methods}

\subsection{EEG Data}
This study used the EEG data recorded and reported in a previous study \cite{nakanishi2019facilitating}. The dataset consisted of the EEG recordings from ten healthy adults.

During the experiment, forty visual stimuli were presented on a 27-inch liquid-crystal display (LCD).
Each stimulus  was  modulated  by  a  sampled-sinusoidal  stimulation method  with  joint  frequency-phase  modulation  (JFPM)  \cite{chen2015high, nakanishi2017enhancing}.  The  stimulation frequencies  ranged  from  8  Hz  to  15.8  Hz with  an  interval of 0.2  Hz.  The  initial  phase values started from 0 rad and the phase interval was 0.35 $\pi$ rad. 

\begin{figure}[t]
      \centering
      \includegraphics[scale=0.52]{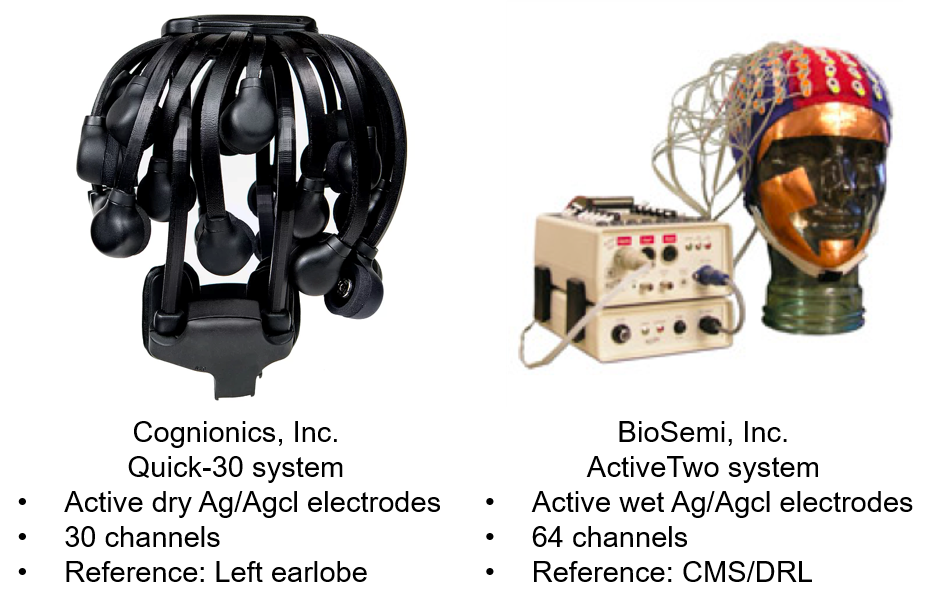}
      \caption{Specifications of two EEG devices used in this study.}
      \label{device}
\end{figure}

In the experiment, the subjects performed two sessions of simulated online experiments \cite{nakanishi2014generating}. The procedure of tasks in both sessions were identical except that the EEG signals were recorded with two different devices. The two devices were an ActiveTwo system (BioSemi, Inc.) as a high-density laboratory-oriented system and a Quick-30 (Q30) system (Cognionics, Inc.) as a mobile and wireless system for real-life applications. Fig. \ref{device} lists the characteristics of the two devices.
The Q30 system was always tested in the first session, and the ActiveTwo system was tested in the second one to avoid the skin preparation required for the wet (gel) electrodes.

In each session, the subjects wore either one of the EEG devices and performed eight blocks of simulations. In each block, the subjects were asked to gaze at one visual stimulus indicated by the stimulus program for 1.5 s at a time until all forty stimuli were gazed once. Therefore, the subjects performed 40 trials corresponding to 40 stimuli in a block, and the data consisted of eight trials per stimulus from each subject. 

\subsection{Preprocessing}

Six channels (PO3, PO4, PO7, PO8, O1, O2) and eight channels (POz, PO3, PO4, PO7, PO8, Oz, O1, O2) of EEG signals were extracted from the recordings collected by the Q30 and the ActiveTwo systems, respectively.
The signals extracted from both devices were resampled at 256 Hz (from 500 Hz and 2048 Hz for the Q30 and the ActiveTwo system respectively) and then re-referenced to the Fz electrode. We employed a conventional sampling-rate conversion method, which includes upsampling, anti-aliasing filtering, and downsampling, to resample the data at a rational ratio between the original and the new sampling rates \cite{crochiere1979general}. This process can be done by using the \textit{resample} function in MATLAB. The re-referencing was done by simply subtracting the signals at target channels by that at the reference channel Fz.

The data were then extracted in [$L$ s, $L$ + 1.5 s], where time zero indicates the stimulus onset and $L$ indicates latency delay in the experimental environment and the human's visual system.
The latency $L$ was set to 0.17 and 0.15 for the Q30 and the ActiveTwo systems, respectively, according to the previous study \cite{nakanishi2019facilitating}. After epoching, the 60-Hz line noise was suppressed by applying an infinite impulse response (IIR) notch filter to each epoch. 

\subsection{TRCA-based SSVEP detection}

TRCA is a data-driven method aiming to find a spatial filter that maximizes the reproducibility in each trial during task periods \cite{nakanishi2017enhancing, tanaka2013task}. The task-related components extracted by the spatial filter obtained by TRCA have been shown to provide better SNR which can significantly improve the performance of training-based SSVEP detections \cite{nakanishi2017enhancing}. In addition, the TRCA-based method could successfully be combined with the filter bank analysis, which decomposes EEG signals into multiple sub-band components so that independent information embedded in the harmonic components can be efficiently extracted \cite{chen2015filter}. 

In the procedure of the TRCA-based method with filter bank analysis, individual calibration data for the $n$-th stimulus are denoted as $\mathbf{x}_n \in \mathbb{R}^{{N_C} \times {N_S} \times {N_T}}$, $n = 1,2,...,N_F$. Here $N_C$  is the number of channels, $N_S$  is the number of sampling points in each trial,  $N_T$ is the number of trials for each stimulus, and $N_F$  is the number of visual stimuli (i.e., 40 in this study). In the training phase, the calibration data are divided into  $N_K$ sub-bands by a filter bank and become $\mathbf{x}^k_n \in \mathbb{R}^{{N_C} \times {N_S} \times {N_T}}$, $k = 1,2,...,N_K$. The $N_K$ was set to five in this study. In the following parts of this paper, the $i$-th trial of stimulus $n$ in sub-band k will be denoted as $\mathbf{x}^k_{n, i}$. Spatial filters in each sub-band are obtained to maximize the sum of the inter-trial covariance after projecting the multi-channel signals into single-channel ones with the spatial filter. Therefore, the goal is finding the channel weights $\mathbf{w}^k_n \in \mathbb{R}^{N_C}$ to maximize the term
\begin{eqnarray}
V^k_n &=& \sum_{i, j \atop {i \neq j}}^{N_T}{\mathrm{Cov}\left((\mathbf{w}^k_n)^T \mathbf{x}^k_{n, i}, (\mathbf{w}^k_n)^T \mathbf{x}^k_{n, j}\right)} \nonumber \\
&=& (\mathbf{w}^k_n)^T \left(\sum_{i, j \atop {i \neq j}}^{N_T}{\mathrm{Cov}\left( \mathbf{x}^k_{n, i}, \mathbf{x}^k_{n, j}\right)} \right) \mathbf{w}^k_n \nonumber \\
&=& (\mathbf{w}^k_n)^T \mathbf{S}^k_n  \mathbf{w}^k_n.
\end{eqnarray}
Here, $\mathbf{S}^k_n$ is the sum of cross-covariance matrices between all pairs of trials of stimulus $n$ in sub-band $k$. To avoid arbitrary scaling with the weights, instead of finding the $\mathbf{w}^k_n$ that maximizes $\mathbf{V}^k_n$, a constraint term is needed:
\begin{eqnarray}
C^k_n &=& \sum_{i}^{N_T}{\mathrm{Var}\left((\mathbf{w}^k_n)^T \mathbf{x}^k_{n, i}\right)} \nonumber \\
&=& (\mathbf{w}^k_n)^T \left(\sum_{i}^{N_T}{\mathrm{Cov}\left( \mathbf{x}^k_{n, i}\right)}\right) \mathbf{w}^k_n \nonumber \\
&=& (\mathbf{w}^k_n)^T \mathbf{Q}^k_n  \mathbf{w}^k_n \nonumber \\
&=& 1.
\end{eqnarray}
Finally, the weights can be calculated as:
\begin{eqnarray}
\mathbf{w}^{k}_n &=& \argmax_{\mathbf{w}} \frac{V^k_n}{C^k_n} \nonumber \\
&=& \argmax_{\mathbf{w}} \frac{\mathbf{w}^T \mathbf{S}^k_n\mathbf{w}}{\mathbf{w}^T \mathbf{Q}^k_n \mathbf{w}}. \label{argmaxw}
\end{eqnarray}
The solution of equation \ref{argmaxw} is equal to the eigenvector of the matrix $\mathbf{Q}^{-1}\mathbf{S}$ with the largest eigvenvalue.

In the ensemble TRCA as an extension version of TRCA, the final spatial filters for each sub-band $\mathbf{w}^k\in \mathbb{R}^{{N_C} \times {N_F}}$ are obtained by concatenating all the weights in each stimulus:
\begin{eqnarray}
    \mathbf{w}^k = \left[ \mathbf{w}^k_1, \mathbf{w}^k_2, \dots, \mathbf{w}^k_{N_F} \right].
\end{eqnarray}
After obtaining the spatial filters, individual templates $\mathbf{\bar{x}}^k_n \in \mathbb{R}^{{N_C} \times {N_S}}$ are prepared. The training trials for $n$-th stimulus in sub-band $k$ are first averaged across trials as:
\begin{eqnarray}
{\mathbf{\bar{x}}}^k_n = \frac{1}{N_T} \sum_{i}^{N_T} \mathbf{x}^k_{n, i}.
\end{eqnarray}

In the testing phase, single-trial testing data $\hat{\mathbf{x}} \in \mathbb{R}^{{N_C} \times {N_S}}$ are first pre-processed by the filter banks to be decomposed into $N_K$ sub-bands as well. Then, the spatial filters $\mathbf{w}^k$ obtained in training phase are applied to the testing signals $\hat{\mathbf{x}}^k \in \mathbb{R}^{{N_C} \times {N_S}}$ in each sub-band. Feature values $\rho^k_n$ are calculated as correlation coefficients between the testing signals and the individual templates as
\begin{eqnarray}
\rho^k_n = r\left((\mathbf{w}^k)^T\mathbf{\hat{x}}^k, (\mathbf{w}^k)^T\mathbf{\bar{x}}^k_n\right),
\end{eqnarray}
where $r\left(\mathbf{a}, \mathbf{b}\right)$ indicates the Pearson’s correlation analysis between two variables $\mathbf{a}$ and $\mathbf{b}$. 
A weighted sum of the ensemble correlation coefficients corresponding to all the sub-bands was calculated as the final feature for target identification as: 
\begin{eqnarray}
\rho_n &=& \sum_{k = 1}^{N_K} \alpha(k) \cdot \rho^k_n,
\end{eqnarray}
where $\alpha(k)$ was defined as $\alpha(k) = k^{-1.25} + 0.25$ according to \cite{chen2015filter}.
Finally, the target stimulus $\tau$ can be identified as \begin{eqnarray}
\tau = \argmax_{n}\rho_n.
\end{eqnarray}

\subsection{LST-based cross-domain transferring}
This work proposes a transformation of SSVEP signals from one domain and another.
Let $\mathbf{x}\in \mathbb{R}^{{N_C} \times {N_S}}$ and $\acute{\mathbf{x}}\in \mathbb{R}^{{N'_C} \times {N_S}}$ be the single-trial SSVEP data obtained in a domain and in another domain (i.e., subject, session, and/or device), respectively.
Then, we aim to find a transformation matrix $\mathbf{P} \in \mathbb{R}^{{N_C} \times {N'_C}}$ such that $\mathbf{x}(t) = \mathbf{P}\acute{\mathbf{x}}(t)+\mathbf{\epsilon}$, where $\mathbf{x}(t), \acute{\mathbf{x}}(t)$ represent the $t$-th column of $\mathbf{x}, \acute{\mathbf{x}}$, and  $\mathbf{\epsilon}\in\mathbb{R}^{N_{C}}$ is an error vector. Note that the numbers of channels $N_C$ and $N'_C$ are not necessary to be equal between two domains.
The transformation matrix $\mathbf{P}$ can be obtained by minimizing the error term $\mathbf{\epsilon}$ in the aforementioned equation with a multivariate least-squares regression given $\mathbf{x}$ and $\acute{\mathbf{x}}$ as follows:
\begin{eqnarray}
\mathbf{P} &=& \argmin_{\mathbf{p}} {\mathrm{Tr}\left[(\mathbf{x} - \mathbf{p} \acute{\mathbf{x}})(\mathbf{x} - \mathbf{p} \acute{\mathbf{x}})^{T}\right]}
\end{eqnarray}
This problem can be solved as follows:
\begin{eqnarray}
\mathbf{P} &=& \mathbf{x} \acute{\mathbf{x}}^T(\acute{\mathbf{x}} \acute{\mathbf{x}}^T)^{-1}
\end{eqnarray}

\begin{figure}[t]
      \centering
      \includegraphics[scale=0.38]{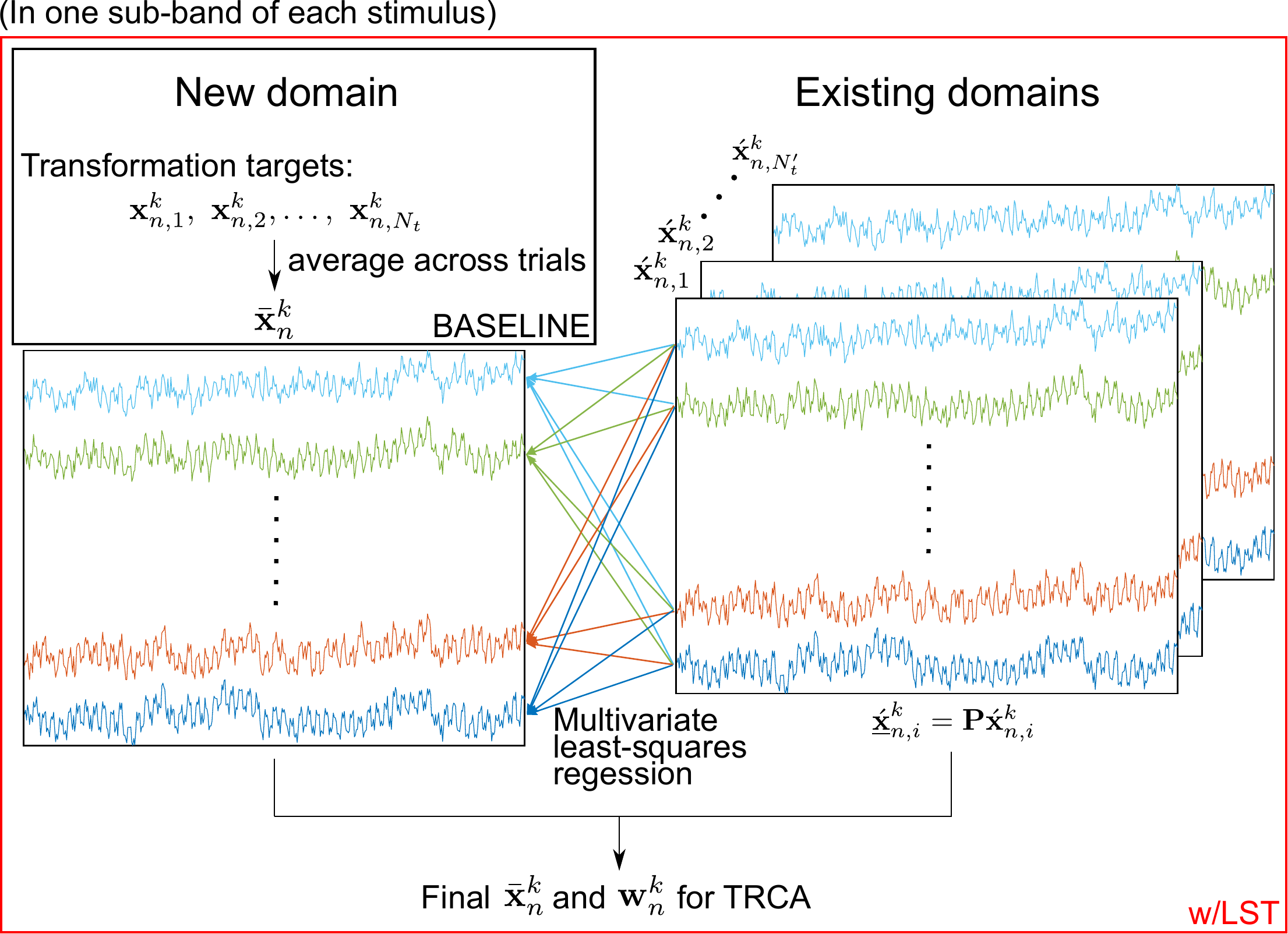}
      \caption{The procedure of transferring SSVEPs based on the least square error transformation. $\mathbf{w}_n^k$ refers to the TRCA-based spatial filer for $n$-th stimulus in $k$-th sub-band (see Equation \ref{argmaxw}).}
      \label{LST}
\end{figure}

Because several studies have shown that trial-averaging can significantly improve the SNR of SSVEPs compared with single-trial SSVEPs by removing background EEG activities \cite{nakanishi2014high,nakanishi2017enhancing}, we used it to improve the transferability of SSVEPs across different domains.
In the proposed method, therefore, instead of using the single-trial signals of the new domain $\mathbf{x}$, we use the averaged signals $\mathbf{\bar{x}}$ across several trials of the signals obtained from the new domain as the target of transformation from existing training pool. These calibration trials obtained from a new user are called transformation targets. Every trial of the existing domains in the training pool will be transformed to signals $\mathbf{\underline{\acute{x}}}$ which should be comparable to the transformation target $\mathbf{\bar{x}}$, i.e. $\mathbf{\bar{x}} \approx \mathbf{\underline{\acute{x}}}_i = \mathbf{P}\acute{\mathbf{x}}_i$ ( $i$ is a trial index). Finally, all trials from the new domain $\mathbf{x}$, which are used to construct the transformation targets, and in the existing data pool $\mathbf{\underline{\acute{x}}}$ are concatenated to form a larger training set than the one used in the conventional template-based algorithm (i.e., $\mathbf{x}$).
With the new training set, the aforementioned TRCA-based method is performed to identify target stimuli. The procedure of the LST is also illustrated in Fig. \ref{LST}.

\begin{figure}[t!]
      \centering
      \includegraphics[scale=0.14]{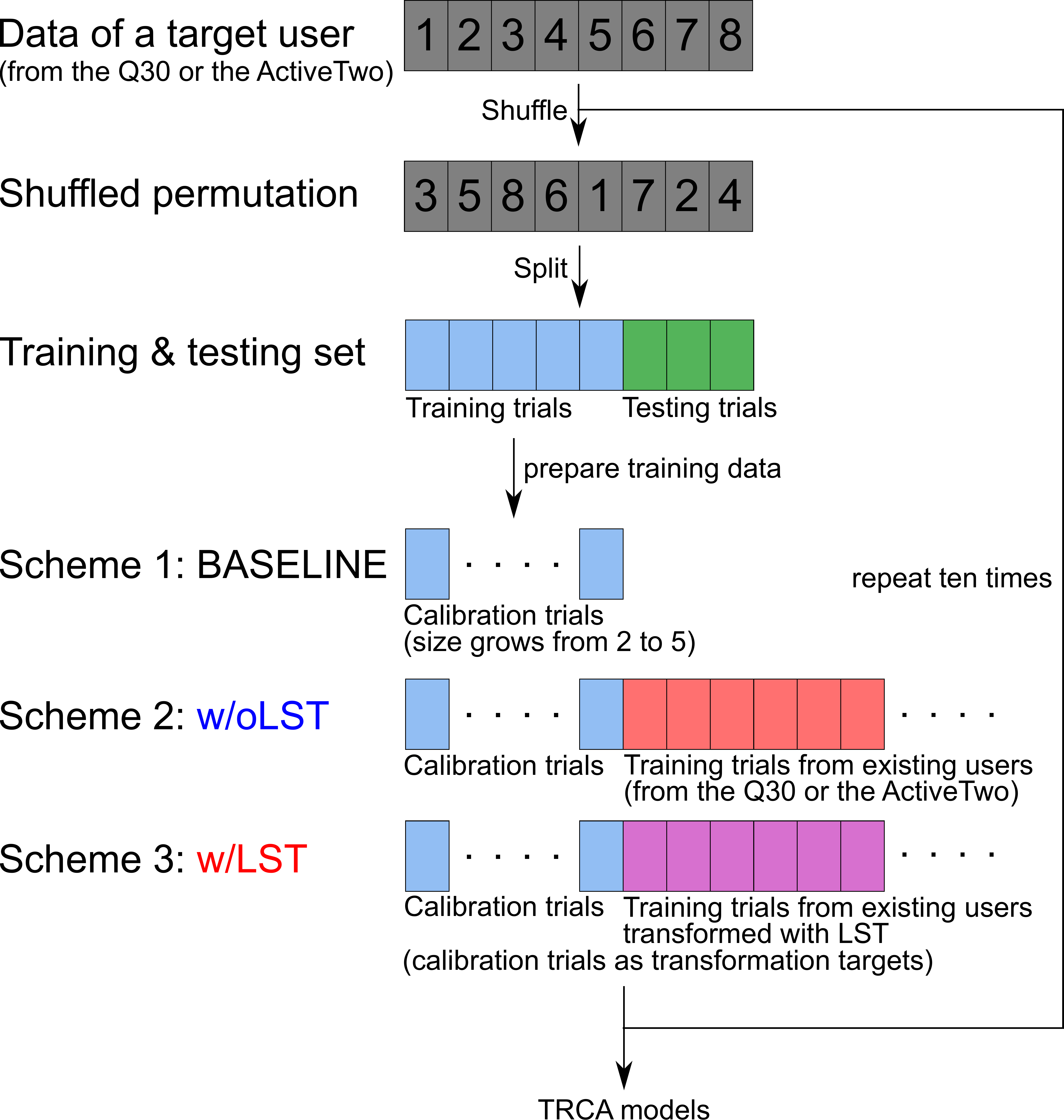}
      \caption{The flowchart of the preparation of the calibration data for three schemes.}
      \label{Flow}
\end{figure}

\begin{figure*}[th!]
      \centering
      \includegraphics[width = 1.0 \textwidth]{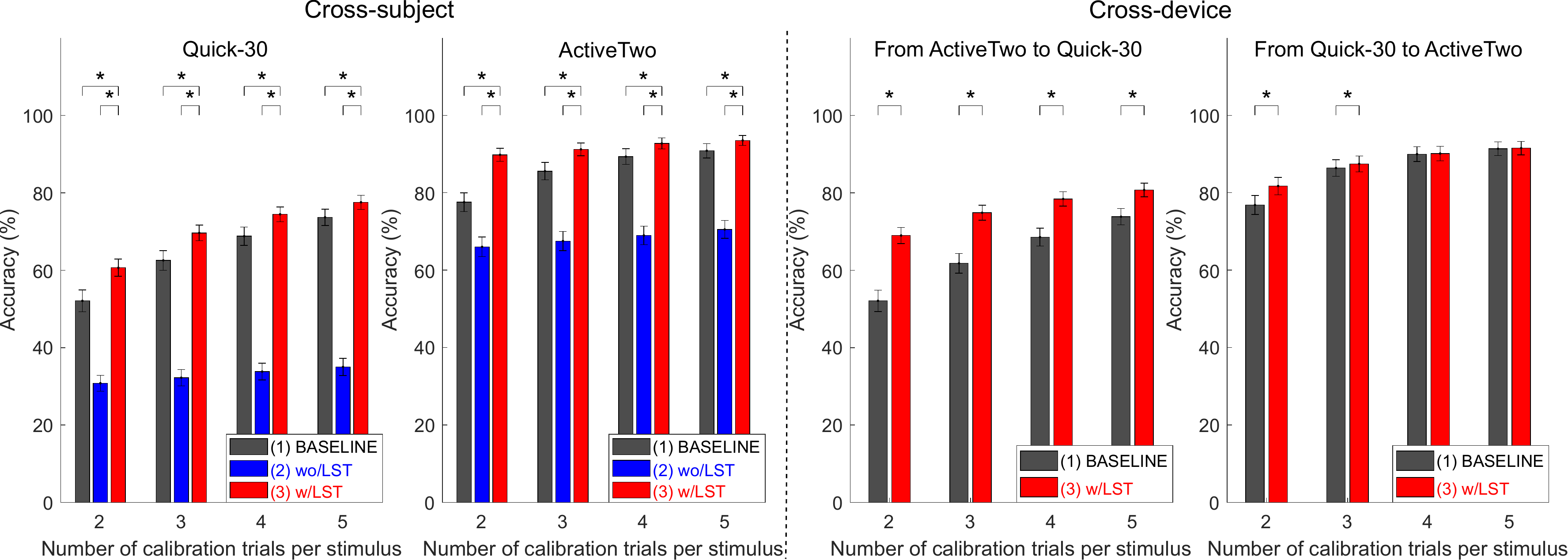}
      \caption{The averaged classification accuracy of different schemes across ten target subjects and ten cross-validation iterations at different numbers of calibration trials per stimulus. '*' indicates $p < 0.05$ of the Wilcoxon signed-rank test between two schemes.}
      \label{bar_plot}
\end{figure*}

\subsection{Performance Evaluation}

To validate the efficacy of the proposed LST-based method in transferring SSVEPs, we compared the performance of detecting SSVEPs using the following three schemes (Fig. \ref{Flow}):
\begin{enumerate}
\item BASELINE: A self-decoding approach in which all the calibration data are collected from a new user (i.e., the conventional individual-template-based method).
\item Transfer without LST (w/oLST): A cross-domain transferring approach in which the calibration data consist of calibration trials from the new domain and from other domains without any transformation.
\item Transfer with LST (w/LST): A cross-domain transferring approach in which the calibration data consist of calibration trials from the new domain and from other domains transformed using LST. The transformation targets are obtained from the data obtained in the new domain.
\end{enumerate}
This study ran a series of simulations to test the performance of the proposed LST-based method as a cross-domain transfer learning for an SSVEP-based BCI.  A leave-one-subject-out cross-validation, in which a subject is treated as a new (i.e., target) user and all the other subjects are treated as existing (i.e., non-target) users, was employed to investigate the effectiveness of the proposed method under the cross-subject scenario. When one session of the new user is being tested, the eight trials for each stimulus was randomly divided into five and three as a calibration set and a testing set. We then trained three models including the TRCA-based spatial filtering and individualized template using different pools of training trials for each scheme.
In the BASELINE, 2-5 calibration trials for each of 40 stimuli from a target subject are used to form training sets. In the w/oLST, all the eight trials for each stimulus from all nine non-target subjects (72 trials in total for each stimulus) are simply merged with the training sets used in the BASELINE. In the w/LST, the data from the non-target subjects are first transformed via the LST and then merged with the training sets used in the BASELINE.
Then, three models were evaluated with SSVEP-decoding performance on the 120-trial (i.e., three trials $\times$ 40 stimuli) testing set. The w/LST scheme was further evaluated under the cross-device scenario. In the scenario, the leave-one-subject-out cross validation was also employed. In addition, different EEG devices were selected between target and non-target subjects. 
Note that the w/oLST cannot be applied to the cross-device scenario because the numbers of electrodes of the two EEG systems are different across devices.
The random separation of the template/test set was repeated ten times. The decoding performance of each target subject was estimated by the average of ten repeats.

Lastly, the classification accuracy was statistically tested by factorial nonparametric permutation-based repeated measures analysis of variances (ANOVAs) \cite{anderson:2003}. The number of permutation was set to 5,000. In the post-hoc analyses, different schemes were compared pair-wisely using Wilcoxon signed-rank tests.

\begin{figure*}[t!]
      \centering
      \includegraphics[width = 1.0 \textwidth]{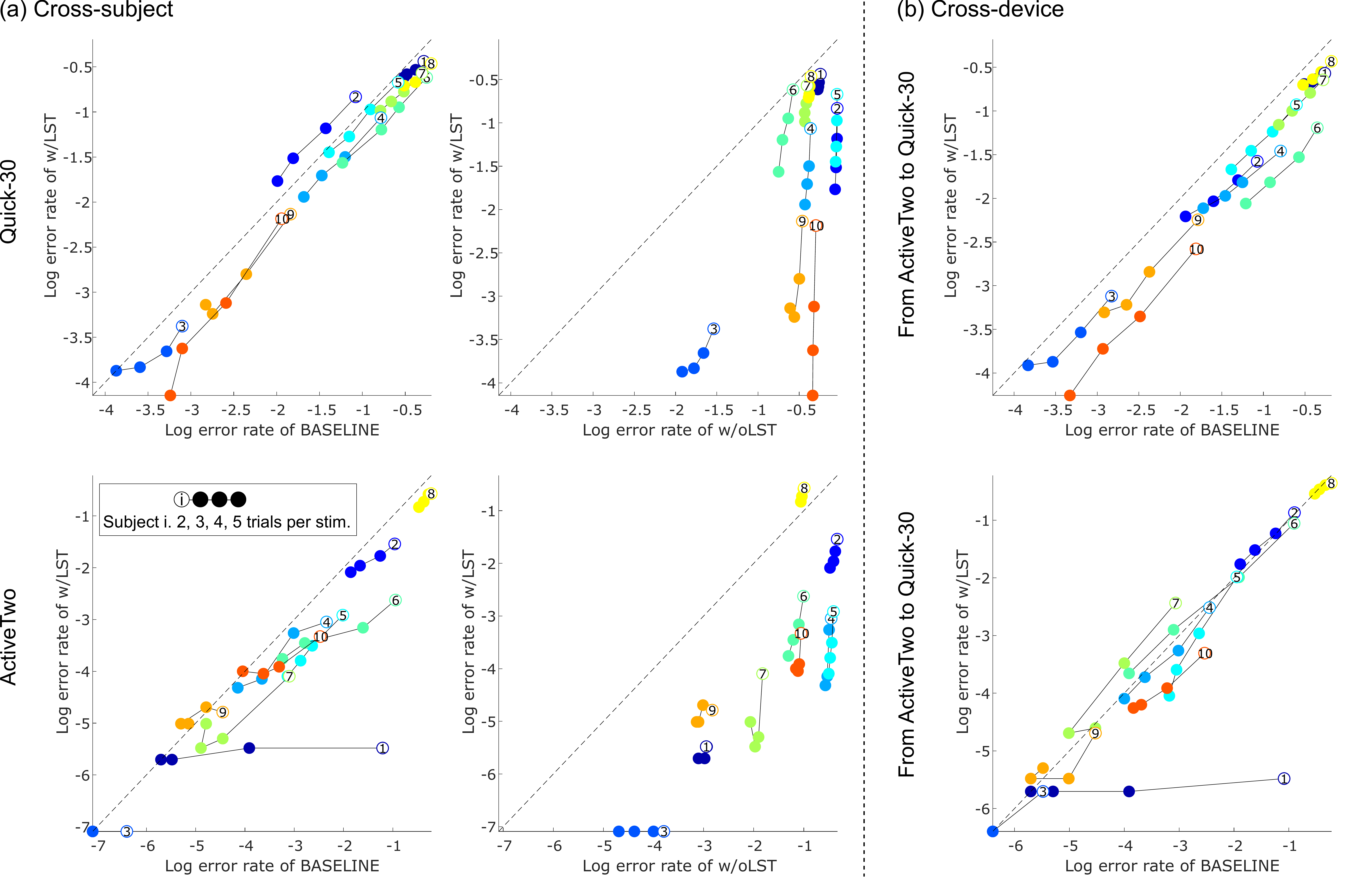}
      \caption{The comparison of the log error rate of different schemes when each of ten subjects plays as the target subject under: (a) cross-subject scenarios, and (b) cross-device scenarios The number in the hollow dots indicates the subject id serving as the target subject. Four dots starting from the hollow dots are cases of number of calibration trials per stimulus starting from two to five.}
      \label{scatter}
\end{figure*}

\begin{figure*}[t!]
      \centering
      \includegraphics[width = 1.0 \textwidth]{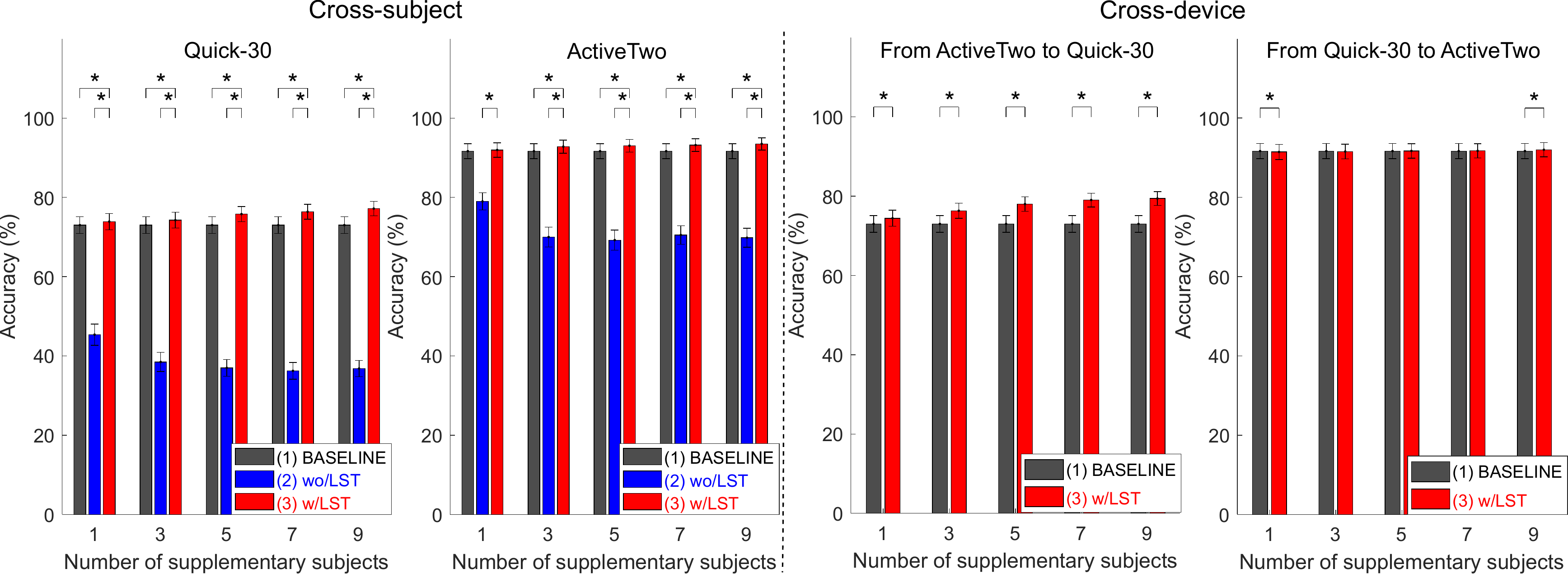}
      \caption{The averaged classification accuracy of different schemes across ten target subjects and ten iterations using different numbers of supplementary subjects. The number of trials per stimulus was fixed to five. '*' indicates $p < 0.05$ of the Wilcoxon signed-rank test between two schemes.}
      \label{bar_plot_subjNum}
\end{figure*}

\section{Results}
Fig. \ref{bar_plot} shows, for the three schemes, the averaged SSVEP-decoding accuracy across subjects with different numbers (from two to five) of calibration trials per stimulus under the cross-subject and cross-device scenarios. In general, the w/LST-based scheme outperformed the other two schemes regardless of the number of calibration trials.

In the cross-subject scenario, a three (schemes) $\times$ four (the number of calibration trials) two-way nonparametric permutation-based repeated measures ANOVA  showed significant main effects of both schemes (Q30: $p < 0.001$; ActiveTwo: $p = 0.006$) and the number of calibration trials (Q30: $p < 0.001$; ActiveTwo: $p < 0.001$). The two-way ANOVA also showed a significant interaction between schemes and the number of calibration trials ($p < 0.001$). In the cross-device scenario, when transferring data from the ActvieTwo to the Q30 systems, a two (schemes) $\times$ four (the numbers of calibration trials per stimulus) two-way ANOVA showed signification main effects of both schemes ($p < 0.001$) and the number of calibration trials ($p < 0.001$), and a significant interaction between them ($p < 0.001$). On the other hand, when transferring from the Q30 to the ActiveTwo system, the two-way ANOVA showed significant main effects of the number of calibration trials ($p < 0.001$), but no significant main effect of schemes ($p = 0.137$) and interaction between them ($p = 0.149$).

The post-hoc Wilcoxon signed-rank tests showed that the w/LST scheme consistently and significantly outperformed the others regardless of the number of calibration trials in the cross-subject scenario. In the cross-device scenario, when transforming the signals from the ActiveTwo system to the Q30 system, the signed-rank tests also showed that the w/LST scheme had significantly higher accuracy than the BASELINE regardless of the number of calibration trials. However, there was no statistically significant difference between w/LST and the BASELINE when transforming signals from the Q30 system to the ActiveTwo system.

Fig. \ref{scatter}(a) and (b) show the decoding performance for each subject under the cross-subject and cross-device scenarios, respectively, with different numbers of calibration trials.
The performance is represented as a logarithmic error rate in the 40-class classification.
In Fig. \ref{scatter}(a) and (b), most of the data points fall within the lower right region (i.e., below the diagonal dashed line), which indicates the w/LST outperformed both the method w/oLST and the BASELINE schemes under most of the circumstances. In particular, when the testing data are from Q30, the w/LST consistently has lower error rates than the BASELINE among nearly all subjects. As for the circumstances when the testing data are from the ActiveTwo system, the w/LST can still outperform the BASELINE among most of the subjects when the size of transformation targets is small. When compared to the method w/oLST, the w/LST has higher accuracy under nearly all circumstances.

\section{Discussions}
\begin{figure}[!htb]
      \centering
      \includegraphics[scale = 0.55]{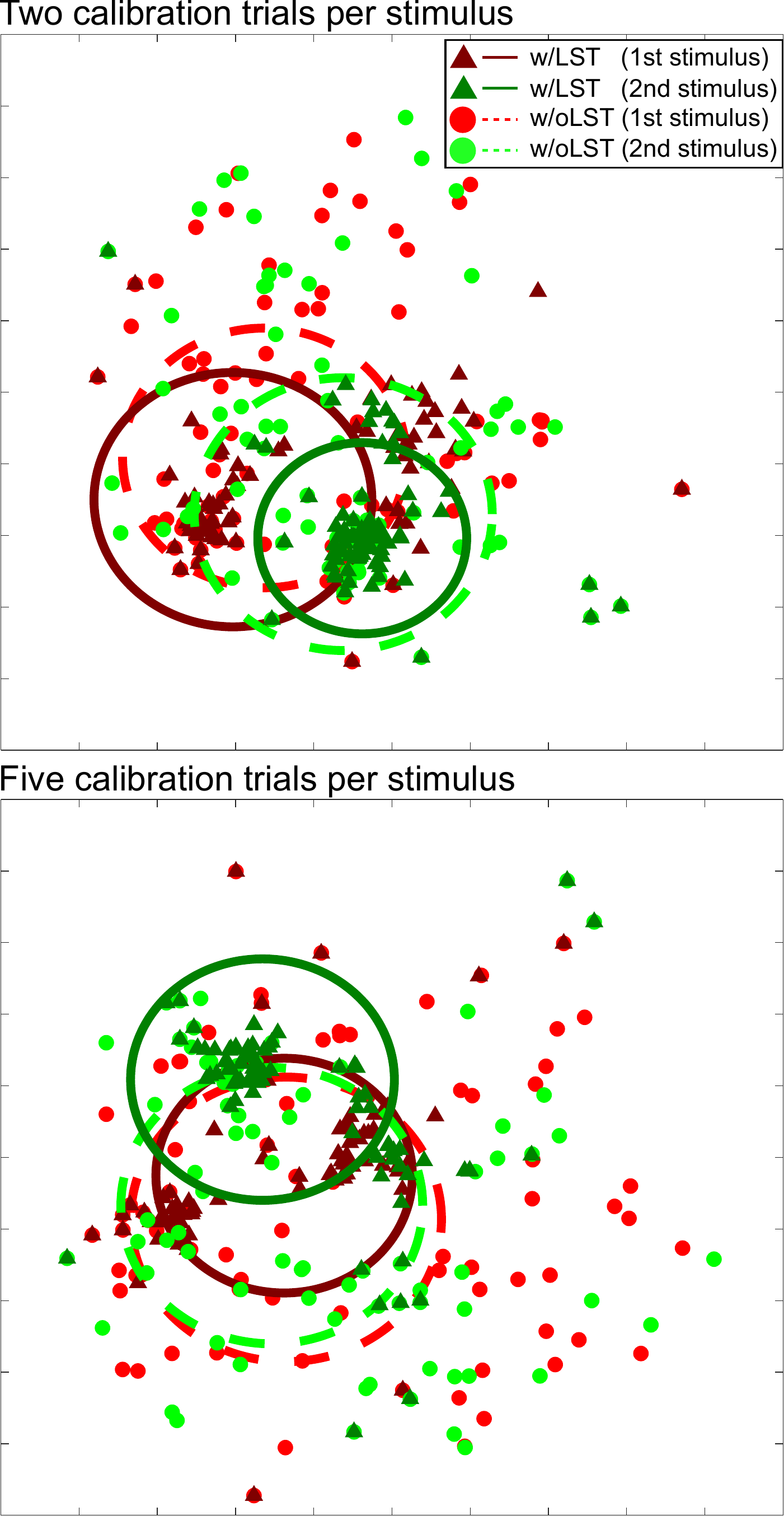}
      \caption{The scatter plot of EEG trials after dimension reduction with t-SNE. For the easiness of visualization, only trials of the first two stimuli are plotted. The triangular dots with darker colors are trials after the LST (subject 1 as the target subject), and the circular dots with lighter colors are original trials. The circles with solid or dashed lines indicate the medians and the standard deviation of trials w/ or w/oLST with corresponding colors. The Silhouette scores of w/oLST and w/LST are 0.0287 and 0.2262 when the number of calibration trials per stimulus is two, and are -0.0071 and 0.1386 when the number of calibration trials per stimulus is five.}
      \label{tsne}
\end{figure}

This study demonstrated the efficacy of the LST-based transfer-learning method in mitigating the variability of SSVEP data across multiple domains. Figs. \ref{bar_plot} and \ref{scatter} suggest that the LST-based method (i.e., w/LST) is capable of boosting the performance of the template-based SSVEP decoding with TRCA especially when the amount of calibration data from the target subject (new user) is insufficient. In addition, study results indicate the negative effects of using the naive transfer learning (w/oLST), compared to the standard TRCA (BASELINE) scheme. These results can be observed from the aspect of the signal characteristics.
Fig. \ref{tsne} shows the scatter plots of sample EEG data recorded with the ActiveTwo system in two schemes (w/ and w/oLST). The subject 1 served as the target subject in both cases, and the sizes of transformation targets were two and five trials per stimulus for the upper and lower plot respectively. The original EEG data and the EEG data transformed by LST onto the transformation targets were pooled together. All EEG trials were first averaged across channels and then processed with t-SNE \cite{maaten2008visualizing} to reduce the dimension to 2D. The plots suggested that the LST can enhance the similarity between trials in the same stimulus across all subjects and reduce similarities across different stimuli. For better visualization, Fig. \ref{tsne} only plots the trials of the first two of 40 stimuli (correspond to two colors, please see the Figure caption).

\begin{figure}[!htb]
      \centering
      \includegraphics[scale = 0.355]{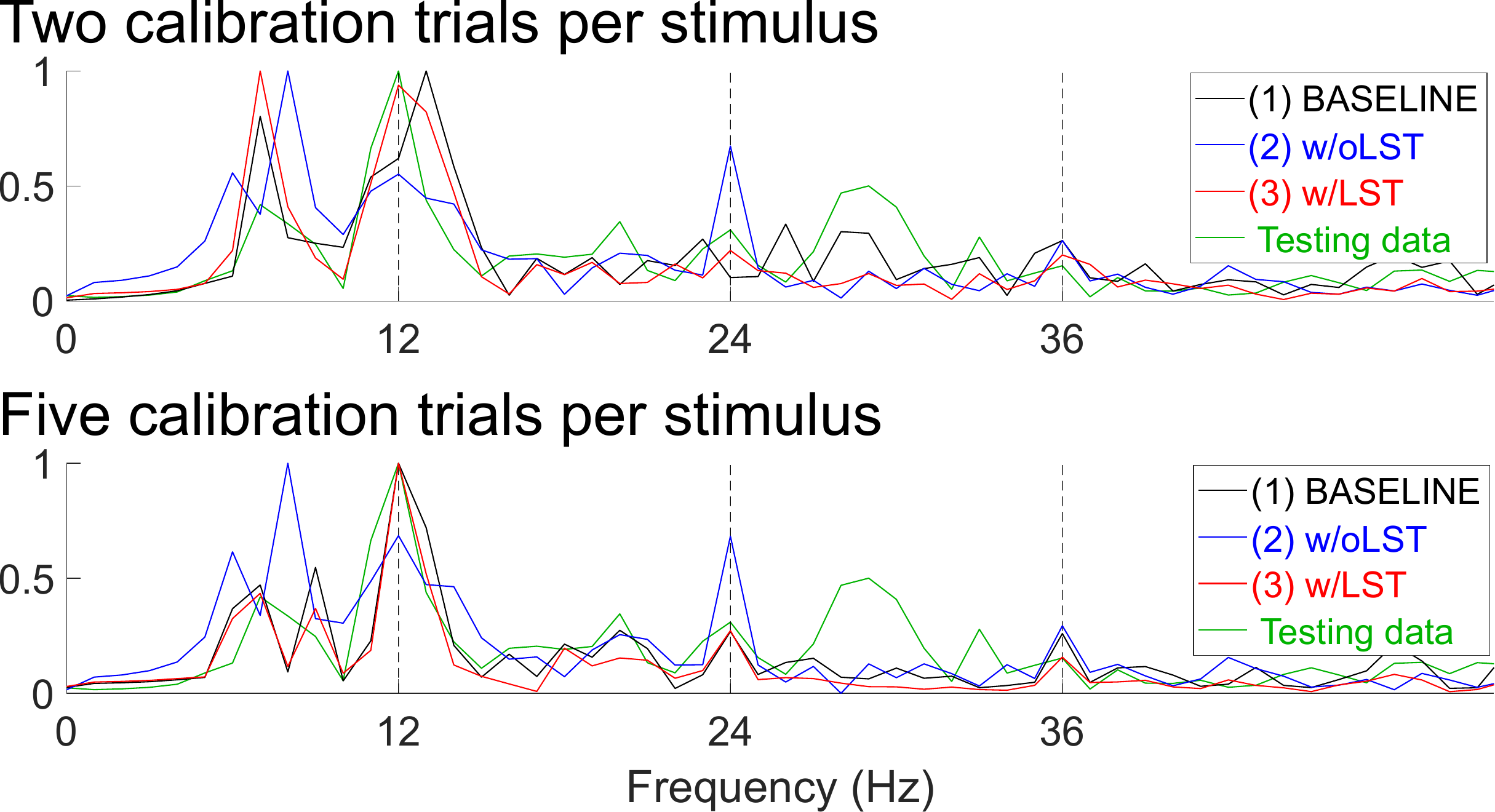}
      \caption{Normalized spectra of averaged EEG signals across all training trials in each scheme (subject 1 as the target subject) and testing trials.}
      \label{spectrum}
\end{figure}

\begin{figure}[t!]
      \centering
      \includegraphics[scale = 0.375]{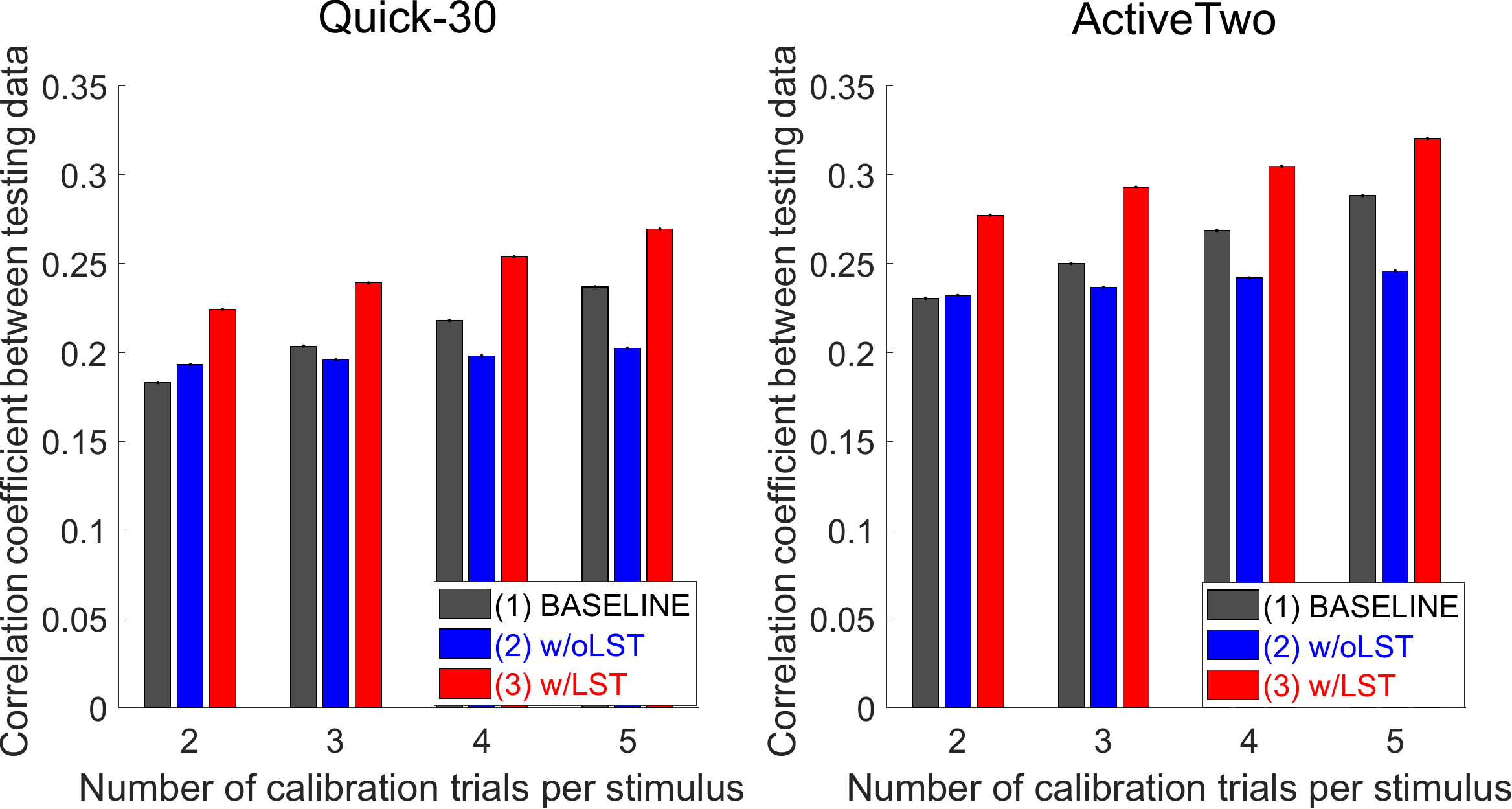}
      \caption{Pearson’s correlation coefficients of time domain signals averaged across channels between training trials and testing trials. The correlation is computed by averaging across all target subjects.}
      \label{timedomain_corrcoef}
\end{figure}

The improvement in the similarity was also reflected in the EEG spectra. Fig. \ref{spectrum} shows the mean spectra of the means across all EEG signals in response to the 12 Hz stimulus in three schemes when subject 1 was the target subject. First, the peak of the spectrum of the BASELINE scheme when the number of calibration trials per stimulus was two (the top panel) didn't appear in the target frequency due to lack of training trials, while the peak became centered at 12 Hz when the number was increased to five (the bottom panel). Note that this phenomenon was reflected in the classification accuracy (Fig. \ref{scatter}b). Furthermore, the fact that the increment in training trials resulted in a more steady spectrum demonstrated the benefit provided by the w/LST scheme. Because the w/LST scheme makes pooling a large number of training trials possible, the SNR can be significantly increased. On the other hand, it can be seen that because the w/oLST scheme simply pooled many trials with high variability, the peak at the target frequency was less prominent. This implies that only with proper transformation on the trials of an existing domain, pooling these trials could lead to positive transfer and improve the SNR. Finally, Fig. \ref{timedomain_corrcoef} shows the averaged Pearson's correlation coefficients of time-domain signals across channels between training and testing trials in all cases. The similarity in the frequency domain matched the magnitude of the correlation coefficients in the time domain. The trend in Fig. \ref{timedomain_corrcoef} also matches the one in the classification accuracy (Fig. \ref{bar_plot}).

The classification results shown in Fig. \ref{bar_plot} and \ref{scatter} suggest the efficacy of the proposed LST-based method, which significantly enhanced SSVEP-decoding performance, particularly when the performance of the original model (BASELINE) was limited. While the leading-edge SSVEP-decoding method, template-based method with TRCA-based spatial filtering \cite{nakanishi2017enhancing}, struggles with time-consuming calibration sessions, the LST-based method can leverage existing data from other domains (subjects and recording devices) and improve decoding performance. As shown in the figures, when the number of template trials was limited in all four scenarios, the w/LST scheme offered high accuracy without requiring many templates.


Comparable performance was found using the conventional TRCA approach (BASELINE) and the w/LST scheme in some cases when testing trials were from the ActiveTwo system (Fig. \ref{scatter}(a), the lower-left panel). In the cross-subject scenarios with the ActiveTwo system, as the number of calibration trials per stimulus was large enough (four and five), the performance of the BASELINE model nearly reached the ceiling, and therefore, the LST-based method couldn't improve the performance, suggesting that leveraging a large amount of data from other subjects has no observable benefit when newly collected individual calibration trials are sufficient. This is in line with the rationale of training-based SSVEP methods, which emphasizes the importance of individualized calibration for SSVEP decoding.

In comparison with the na\"ive  transferring (w/oLST), for most of the subjects, the performance of the LST-based method improved along with acquiring additional calibration trials from the target user while the w/oLST scheme didn't. In addition, another big advantage of the LST is that the numbers and locations of the EEG channels of new calibration data and supplementary data can be different. In the cross-device scenario, while the na\"ive data pooling is not even allowed, the LST could help expand the number of training trials.

In the cross-device scenario, when the EEG signals of the target subjects were from the ActiveTwo system and the ones of the existing subjects were from the Q30 system, the increment in accuracy that the LST can bring was less than transferring within the ActiveTwo system (Fig. \ref{bar_plot}, the second and the fourth panel from the left).
This implies the limitation of the LST-based method that it still relies on the supplementary data from existing domains with good quality. Nonetheless, the LST did not bring any negative impact either. In a more practical scenario, in which the dry-EEG-based Q30 was used as the recording device for a new target subject, the LST can leverage the existing data collected from other subjects using a standard wet-electrode-based EEG system in a well-controlled laboratory to improve the SSVEP-decoding accuracy (Fig. \ref{bar_plot}, the third panel from the left). In other words, if there is sufficient data collected by gel-based EEG systems in well-controlled laboratories or even from the publicly available datasets on the Internet, the LST can leverage these data and a small number of calibration data collected from the test subject to build a practical SSVEP BCI, significantly improving the practicality of real-world SSVEP BCIs.

In the two scenarios, the cross-subject transferring within ActiveTwo system and the cross-device transferring from the ActiveTwo to the Q30 system, the accuracy of the w/LST with two calibration trials per stimulus was equal to or higher than that of the BASELINE with five calibration trials per stimulus. Therefore, in such cases, 60\% of calibration time could be saved. Assuming a 40-command BCI speller, the proposed method could save five minutes to collect training data with a trial length of 2.5 s (Stimulation time: 1.5 s; Inter-stimulus interval: 1.0 s; Three trials per stimulus).

In addition to the simulation validating the LST-based method with different numbers of calibration trials, another simulation study that varied the number of supplementary subjects was also conducted. In this simulation, a leave-one-subject-out cross-validation was also employed, but when preparing the calibration data of the w/oLST and w/LST schemes, different numbers (1, 3, 5, 7, and 9) of other subjects were randomly selected as the supplementary subjects. For each number of supplementary subject, the random selections were repeated ten times. Unlike the first simulation, the number of trials per stimulus was fixed to five, and the first five trials were always used to form the calibration data and the last three rials were used as testing trials. 
As Fig. \ref{bar_plot_subjNum} shows, the performance of the w/LST scheme method improved slightly as the number of supplementary subjects increased, while the w/oLST scheme did not. These results suggested that the number of supplementary subjects does not heavily affect the performance of the LST-based method but the more the better. These results are important for the practical scenarios, because in real-world, the number of supplementary subjects (i.e., existing users) is not limited, and it's important to show that the LST-based method doesn't rely on very specific parameters.

In a comparison of the proposed method with existing approaches listed in Table \ref{table_compare}, our work stands out as the only generalized framework for multiple cross-domain transfer learning for SSVEP decoding. Although some of the methods do not require any additional calibration data from a target session, the others require a small amount of calibration data from a target session. In general, the methods that do not require additional data can achieve higher accuracy than the calibration-free method, but they are far inferior to fully-calibrate methods such as the TRCA-based method \cite{nakanishi2019facilitating}. The ones that require calibration data including the LST-based method employ transfer learning to achieve better performance than fully-calibration methods. Most importantly, most of the existing transfer learning methods can only transfer SSVEPs across one domain such as either cross-subject, cross-session or cross-device scenarios, whereas this study validated that the LST-based method can transfer SSVEPs across multiple domains except cross-stimulus transferring. It indicates that any user could reach a higher accuracy than fully-calibrated methods with a small amount of calibration data collected from his/her EEG device even if it was the first time for him/her to use the system.

It is also worth noting that the classification accuracy obtained in the simulated online analysis could be generalized to actual online performance. The preprocessing pipeline including notch and band-pass filters was applied to each data epoch separately after the data were segmented. In addition, the inter-stimulus interval (ISI) was set to 1.0 s in the experiment, which has been commonly and reasonably used in previous studies with online analyses \mbox{\cite{chen2015high, nakanishi2017enhancing}}.

The main limitation of the proposed method is that it still requires a small number of calibration trials from the new user, and therefore, it's still yet a calibration-free method. In addition, when the quality of the supplemental data is worse than the targeting data, the improvement from the LST-based method is limited. However, in practice, it's more likely that the supplemental data have better signal quality since they can be prepared in a well-controlled environment while the data of the targeting user can be acquired in any general environment.

In a nutshell, the LST enables an effective consolidation of EEG data across users and devices and consistently outperforms the standard TRCA approach and the naive integration of data without LST. Our results suggest using the LST-based method should be taken into account for augmenting calibration data when using TRCA-based SSVEP spellers.

\section{Conclusions}
This study proposed a cross-domain transfer method based on the LST for transforming SSVEP data across users and devices. The experimental results demonstrated the efficacy of the LST-based method in alleviating the inter-subject and inter-device variability in the SSVEPs. The LST-based method also improved the SSVEP-decoding accuracy by leveraging data from other subjects and/or devices and a small number of calibration data from a new subject. These findings considerably improve the practicality of a real-world SSVEP BCI.

\section*{Acknowledgement}
This work was partially supported by the US Army Research Laboratory under Cooperative Agreement W911NF-10-2-0022. T-P Jung is also supported, in part, by the US National Science Foundation (NSF) under Grant CBET-1935860. This work was also supported in part by the Higher Education Sprout Project of the National Chiao Tung University and Ministry of Education, Taiwan, the Ministry of Science and Technology, Taiwan, under MOST(NSC) 109-2222-E-009-006-MY3.

\section*{References}
\bibliographystyle{unsrt.bst}
\bibliography{ref}

\begin{thebibliography}{10}

\bibitem{wolpaw2002brain}
Jonathan~R Wolpaw, Niels Birbaumer, Dennis~J McFarland, Gert Pfurtscheller, and
  Theresa~M Vaughan.
\newblock Brain--computer interfaces for communication and control.
\newblock {\em Clinical neurophysiology}, 113(6):767--791, 2002.

\bibitem{donchin:1988}
L~A Farwell and E~Donchin.
\newblock {Talking off the top of your head: toward a mental prosthesis
  utilizing event-related brain potentials}.
\newblock {\em Electroencephalogr. Clin. Neurophysiol.}, 70(6):510--23, Dec.
  1988.

\bibitem{wang2008brain}
Yijun Wang, Xiaorong Gao, Bo~Hong, Chuan Jia, and Shangkai Gao.
\newblock Brain-computer interfaces based on visual evoked potentials.
\newblock {\em IEEE Eng. Med. Biol. Mag.}, 27(5):64--71, 2008.

\bibitem{vialatte:2010}
Fran\c{c}ois-Beno\^{\i}t Vialatte, Monique Maurice, Justin Dauwels, and Andrzej
  Cichocki.
\newblock {Steady-state visually evoked potentials: focus on essential
  paradigms and future perspectives}.
\newblock {\em Prog. Neurobiol.}, 90(4):418--438, April 2010.

\bibitem{gao2014visual}
S~Gao, Y~Wang, X~Gao, and B~Hong.
\newblock Visual and auditory brain--computer interfaces.
\newblock {\em IEEE Trans. Biomed. Eng.}, 61(5):1436--1447, 2014.

\bibitem{chen2015high}
Xiaogang Chen, Yijun Wang, Masaki Nakanishi, Xiaorong Gao, Tzyy-Ping Jung, and
  Shangkai Gao.
\newblock High-speed spelling with a noninvasive brain--computer interface.
\newblock {\em Proc. Nat. Acad. Sci. U. S. A.}, 112(44):E6058--E6067, 2015.

\bibitem{lin2006frequency}
Zhonglin Lin, Changshui Zhang, Wei Wu, and Xiaorong Gao.
\newblock Frequency recognition based on canonical correlation analysis for
  ssvep-based bcis.
\newblock {\em IEEE Trans. Biomed. Eng.}, 53(12):2610--2614, 2006.

\bibitem{bin:2009b}
Guangyu Bin, Xiaorong Gao, Zheng Yan, Bo~Hong, and Shangkai Gao.
\newblock {An online multi-channel SSVEP-based brain-computer interface using a
  canonical correlation analysis method}.
\newblock {\em J. Neural Eng.}, 6(4):046002, August 2009.

\bibitem{wang2015computational}
Yijun Wang, Xiaorong Gao, and Shangkai Gao.
\newblock {Computational modeling and application of steady-state visual evoked
  potentials in brain-computer interfaces}.
\newblock {\em Sci. Suppl.}, 350(6256):43--46, 2015.

\bibitem{wei2018subject}
Chun-Shu Wei, Yuan-Pin Lin, Yu-Te Wang, Chin-Teng Lin, and Tzyy-Ping Jung.
\newblock A subject-transfer framework for obviating inter-and intra-subject
  variability in eeg-based drowsiness detection.
\newblock {\em NeuroImage}, 174:407--419, 2018.

\bibitem{nakanishi2014high}
Masaki Nakanishi, Yijun Wang, Yu-Te Wang, Yasue Mitsukura, and Tzyy-Ping Jung.
\newblock A high-speed brain speller using steady-state visual evoked
  potentials.
\newblock {\em Int. J Neural Syst.}, 24(06):1450019, 2014.

\bibitem{nakanishi2017enhancing}
Masaki Nakanishi, Yijun Wang, Xiaogang Chen, Yu-Te Wang, Xiaorong Gao, and
  Tzyy-Ping Jung.
\newblock Enhancing detection of ssveps for a high-speed brain speller using
  task-related component analysis.
\newblock {\em IEEE Trans. Biomed. Eng.}, 65(1):104--112, 2018.

\bibitem{nakanishi2015comparison}
Masaki Nakanishi, Yijun Wang, Yu-Te Wang, and Tzyy-Ping Jung.
\newblock A comparison study of canonical correlation analysis based methods
  for detecting steady-state visual evoked potentials.
\newblock {\em PLoS One}, 10(10):e140703, Oct. 2015.

\bibitem{zhang:2013l1regularized}
Yu~Zhang, Guoxu Zhou, Jing Jin, M~Wang, Xing Wang, and Andrzej Cichocki.
\newblock {L1-Regularized Multiway Canonical Correlation Analysis for
  SSVEP-Based BCI}.
\newblock {\em IEEE Trans. Neural Syst. Rehabil. Eng.}, 21(6):887--896, October
  2013.

\bibitem{zhang2014frequency}
Yu~Zhang, Guoxu Zhou, Jing Jin, Xingyu Wang, and Andrzej Cichocki.
\newblock Frequency recognition in {SSVEP}-based {}bci using multiset canonical
  correlation analysis.
\newblock {\em Int, J. Neural Syst.}, 24(04):1450013, 2014.

\bibitem{zerafa2018to}
R.~Zerafa, Tracey Camilleri, Owen Falzon, and Kenneth~P. Camilleri.
\newblock {To train or not to train? A survey on training of feature extraction
  methods for SSVEP-based BCIs}.
\newblock {\em J. Neural Eng.}, 15:051001, Jul. 2018.

\bibitem{Wu2020transfer}
D~Wu, Y~Xu, and B.-L Lu.
\newblock Transfer learning for {EEG}-based brain-computer interfaces: {A}
  review of progress made since 2016.
\newblock {\em IEEE Trans. Cog. Dev. Syst.}, In press.

\bibitem{yuan2015enhancing}
Peng Yuan, Xiaogang Chen, Yijun Wang, Xiaorong Gao, and Shangkai Gao.
\newblock Enhancing performances of ssvep-based brain--computer interfaces via
  exploiting inter-subject information.
\newblock {\em J Neural Eng.}, 12(4):046006, 2015.

\bibitem{wong2020inter}
Chi~Man Wong, Ze~Wang, Boyu Wang, Ka~Fai Lao, Agostinho Rosa, Peng Xu,
  Tzyy-Ping Jung, CL~Philip Chen, and Feng Wan.
\newblock Inter-and intra-subject transfer reduces calibration effort for
  high-speed {SSVEP}-based {BCIs}.
\newblock {\em IEEE Trans. Neural Syst. Rehabil. Eng.}, 2020.

\bibitem{Rodrigues2019-yl}
P~L~C Rodrigues, C~Jutten, and M~Congedo.
\newblock Riemannian procrustes analysis: Transfer learning for
  {Brain--Computer} interfaces.
\newblock {\em IEEE Trans. Biomed. Eng.}, 66(8):2390--2401, August 2019.

\bibitem{Waytowich2018-ph}
Nicholas Waytowich, Vernon~J Lawhern, Javier~O Garcia, Jennifer Cummings, Josef
  Faller, Paul Sajda, and Jean~M Vettel.
\newblock Compact convolutional neural networks for classification of
  asynchronous steady-state visual evoked potentials.
\newblock {\em J. Neural Eng.}, 15(6):066031, December 2018.

\bibitem{Nakanishi2016-qa}
M~Nakanishi, Y~Wang, and T~P Jung.
\newblock Session-to-session transfer in detecting steady-state visual evoked
  potentials with individual training data.
\newblock {\em International Conference on Augmented}, 2016.

\bibitem{nakanishi2019facilitating}
Masaki Nakanishi, Yu-Te Wang, Chun-Shu Wei, Kuan-Jung Chiang, and Tzyy-Ping
  Jung.
\newblock Facilitating calibration in high-speed bci spellers via leveraging
  cross-device shared latent responses.
\newblock {\em IEEE Trans. Biomed. Eng.}, 2019.

\bibitem{Suefusa2017-qp}
K~Suefusa and T~Tanaka.
\newblock Reduced calibration by efficient transformation of templates for high
  speed hybrid coded {SSVEP} brain-computer interfaces.
\newblock In {\em 2017 {IEEE} International Conference on Acoustics, Speech and
  Signal Processing ({ICASSP})}, pages 929--933, March 2017.

\bibitem{chiang2019cross}
Kuan-Jung Chiang, Chun-Shu Wei, Masaki Nakanishi, and Tzyy-Ping Jung.
\newblock Cross-subject transfer learning improves the practicality of
  real-world applications of brain-computer interfaces.
\newblock In {\em Proc. 9th Int. IEEE/EMBS Neural Eng. Conf.}, pages 424--427,
  2019.

\bibitem{nakanishi2014generating}
Masaki Nakanishi, Yijun Wang, Yu-Te Wang, Yasue Mitsukura, and Tzyy-Ping Jung.
\newblock Generating visual flickers for eliciting robust steady-state visual
  evoked potentials at flexible frequencies using monitor refresh rate.
\newblock {\em PLoS One}, 9(6):e99235, 2014.

\bibitem{crochiere1979general}
RE~Crochiere.
\newblock A general program to perform sampling rate conversion of data by
  rational ratios.
\newblock {\em Programs for digital signal processing}, 1979.

\bibitem{tanaka2013task}
H~Tanaka, T~Katura, and H~Sato.
\newblock Task-related component analysis for functional neuroimaging and
  application to near-infrared spectroscopy data.
\newblock {\em NeuroImage}, 64:308--327, 2013.

\bibitem{chen2015filter}
Xiaogang Chen, Yijun Wang, Shangkai Gao, Tzyy-Ping Jung, and Xiaorong Gao.
\newblock Filter bank canonical correlation analysis for implementing a
  high-speed ssvep-based brain--computer interface.
\newblock {\em J. Neural Eng.}, 12(4):046008, 2015.

\bibitem{anderson:2003}
M.~Anderson and C.~Ter Braak.
\newblock {Permutation tests for multi-factorial analysis of variance}.
\newblock {\em J. Stat. Comput. Simul.}, 73(2):85--113, Oct. 2003.

\bibitem{maaten2008visualizing}
Laurens van~der Maaten and Geoffrey Hinton.
\newblock Visualizing data using t-sne.
\newblock {\em J. Mach. Learn. Res.}, 9(Nov):2579--2605, 2008.

\end{thebibliography}

\end{document}